# Mitigating Backdoors within Deep Neural Networks in Data-limited Configuration

Soroush Hashemifar, Saeed Parsa*, Morteza Zakeri-Nasrabadi

hashemifar_soroush@cmps2.iust.ac.ir, parsa@iust.ac.ir, morteza_zakeri@comp.iust.ac.ir

School of Computer Engineering, Iran University of Science and Technology, Tehran, Iran.

*Abstract*—As the capacity of deep neural networks (DNNs) increases, their need for huge amounts of data significantly grows. A common practice is to outsource the training process or collect more data over the Internet, which introduces the risks of a backdoored DNN. A backdoored DNN shows normal behavior on clean data while behaving maliciously once a trigger is injected into a sample at the test time. In such cases, the defender faces multiple difficulties. First, the available clean dataset may not be sufficient for fine-tuning and recovering the backdoored DNN. Second, it is impossible to recover the trigger in many real-world applications without information about it. In this paper, we formulate some characteristics of poisoned neurons. This backdoor suspiciousness score can rank network neurons according to their activation values, weights, and their relationship with other neurons in the same layer. Our experiments indicate the proposed method decreases the chance of attacks being successful by more than 50% with a tiny clean dataset, *i.e.*, ten clean samples for the CIFAR-10 dataset, without significantly deteriorating the model's performance. Moreover, the proposed method runs three times as fast as baselines.

## I. INTRODUCTION

Deep neural networks (DNNs) have gained much attention in various applications. A common practice to train large DNNs is using datasets available over the Internet or pre-trained models from models' zoo platforms. This practice brings a menace to DNNs, known as train-time attacks [1], [2] and third-party attacks [3]. Among them, backdoor attacks [4]–[9] are more popular due to their simplicity for real-world applications. Intuitively, backdoor attacks involve injecting a specifically designed trigger pattern into a small portion of the training dataset. Hence, the DNN learns to establish a strong connection between the trigger pattern and a target label, which gives the attacker full control over the behavior of the DNN.

Considering the hardship of removing the complete effects of a backdoor attack from a DNN, many approaches have been proposed to defend the poisoned DNN [10], [11]. Some research works focus on purifying malicious input [12]–[14], while others aim to cleanse poisoned computational elements, *i.e.*, neurons [15]. This paper focuses on a realistic scenario in which the defender aims to mitigate the backdoor effects without any control over the training process or knowledge of trigger patterns. Some works purify the compromised DNN by fine-tuning [16], [17] and knowledge distillation [18]. Regarding fine-tuning and knowledge distillation techniques, it is hard to mitigate the backdoor effects completely. In addition, the defender has limited access to clean samples in most real-world scenarios. Hence, fine-tuning and distilling the knowledge may end up overfitting and destroying the performance of DNN on clean test data. Other works aim at mitigating the backdoor effects by circumventing fine-tuning complexity through adversarial perturbations [19], [20] and pruning outlier neurons based on their weights' maximum singular value [21]. These works aim at detecting sensitive weights and neurons related to the backdoor and pruning them. However, perturbing neurons still need sufficient available data to identify those sensitive neurons accurately. Moreover, evading data makes it challenging to locate poisoned neurons accurately because data is the key to revealing backdoored behavior. For these reasons, the previously proposed approaches face practical limitations in real-world scenarios.

This paper proposes a lightweight backdoor removal method that prunes the poisoned filters of convolutional layers in a data-limited setting. We define a formulation to measure the backdoor suspiciousness of each filter based on the characteristics of their associated weights and feature maps. Filters with suspiciousness scores above an empirically calculated threshold are then pruned to mitigate the backdoor behavior. Extensive experiments on the CIFAR-10 dataset against four popular backdoor attacks indicate that our proposed method outperforms state-of-the-art trojan repairing techniques regarding attack success rate (ASR) and models' performance degradation rate. The proposed method reduces ASR by more than 50% while degrading the accuracy on clean samples (ACC) by 3 to 7% compared to the original model. Moreover, the experiments exhibit the superior run-time performance of the proposed method compared to baselines involving

---

adversarial perturbation. The main contributions of this paper are summarized as follows:
i. We propose a lightweight yet efficient and effective backdoor mitigation algorithm for data-limited scenarios.
ii. We theoretically formulate the characteristics of poisoned filters in a compromised neural network.
iii. We empirically show the superior performance of the proposed algorithm in removing backdoored filters on the CIFAR-10 dataset against four common backdoor attacks.

The rest of the paper is organized as follows. Section 2 provides related work about various defense approaches against backdoor attacks. Section 3 presents the formulation of the backdoor suspiciousness score and our proposed algorithm to prune poisoned filters. Section 4 discusses the attack scenarios, baselines' configurations, and empirical experimental results. Finally, Section 5 concludes the paper.

## II. RELATED WORK

Backdoor removal techniques aim to eliminate hidden vulnerabilities from compromised models while minimizing any substantial degradation in their performance on clean data samples. A specific line of research involves purifying malicious input samples. Gao et al. [13] proposed a strong intentional perturbation (STRIP) technique to detect backdoor attacks at inference time. STRIP identifies poisoned input samples by superimposing them with a clean set of images and examining the entropy imposed in their predictions. By assuming that the trigger pattern is input-agnostic, smaller randomness in predictions leads to higher probability that the input image is poisoned. In contrast, Udeshi et al. [14] proposed a method, called Neo, which search for a candidate region, where modifying results in drastic change in predictions. In its preprocessing phase, Neo uses the dominant color in the image to create a squarelike trigger blocker, which was chosen to locate and remove the backdoor trigger. Neo assumes that the placement of the trigger blocker in the position of the trigger pattern in the attacked image would significantly alter model predictions. After that, Liu et al. [12] proposed a method, called TrojDef, which investigates the impact of adding Gaussian noise on leading backdoored models to abnormal behavior on trigger samples.

Regarding model repair methods, a line of research aims to refine the compromised model through a process known as fine-tuning. In this context, Li et al. [18] have introduced a refinement procedure called neural attention distillation (NAD), which leverages distillation-guided fine-tuning to enhance the defense of DNNs against backdoor attacks. NAD employs a dual-network framework comprising a teacher network derived from conventional fine-tuning of the compromised model using a large set of clean images, including at least one percent of the training set. For example, researchers have used 500 clean images from the CIFAR-10 dataset to mitigate the backdoor attack in compromised convolutional neural networks. The teacher network serves as a guide for fine-tuning a backdoored student network, operating on a limited subset of clean training data. The objective is to align the intermediate-layer attention patterns of the student network with those of the teacher network, thus enhancing the model's resistance to backdoor threats. The most significant disadvantage of fine-tuning-based methods is that it is hard for them to completely remove the backdoor footprints from the compromised network due to their need for a huge amount of data. Moreover, fine-tuning-based methods tend to overfit or underfit if the number of training samples is insufficient, which is a known issue named *catastrophic forgetting* [22].

Another direction of research on backdoor removal combines pruning and fine-tuning techniques in models that have been compromised. Liu et al. [16] introduced a strategy that pairs pruning and fine-tuning as a natural defense. This method aims to shrink the size of the compromised network by removing inactive neurons that have no impact on clean inputs, thereby deactivating the backdoor functionality. Similarly, Guan et al. [17] proposed the Shapley Pruning framework for identifying and mitigating backdoor attacks originating from manipulated models. This framework considers the relationships between neurons, pinpointing the limited number of affected neurons. It maintains the model's structure and accuracy while simultaneously pruning as many infected neurons as possible. These works suffer from a similar problem to the previously mentioned research line.

Finally, the recent research addresses the need to remove backdoors without resorting to extensive fine-tuning. [19] introduced the adversarial neuron pruning (ANP) technique, which refines the process of pruning backdoored neurons with higher precision. ANP leverages adversarial perturbations on the network's weights to identify and eliminate neurons that exhibit higher sensitivity to adversarial perturbations, signifying their involvement in backdoor behavior. In a related vein, [20] presented the adversarial weight masking (AWM) method, capable of erasing neuron-based backdoors even within a one-shot setting. AWM adopts a mini-max approach to mask parameters in the network adversarially. The essence of AWM lies in reducing weights on parameters linked to backdoor triggers while emphasizing robust features. ANP and AWM fail to locate backdoored neurons in data-limited scenarios since they need sufficient data to detect poisoned neurons accurately. Another approach proposed by Zheng et al. [21], named channel lipschitzness pruning (CLP), focuses on removing channels with higher Lipschitz constants to restore the model. The Lipschitz constant is evaluated for each channel, gauging its responsiveness to inputs. Given that channels detecting backdoor triggers should be particularly sensitive to specific perturbations, the authors contend that channels associated with backdoors should exhibit higher Lipschitz constants than normal channels. The disadvantage of CLP is that it neglects data characteristics, while data plays a crucial role in revealing the backdoor behavior of a compromised DNN.

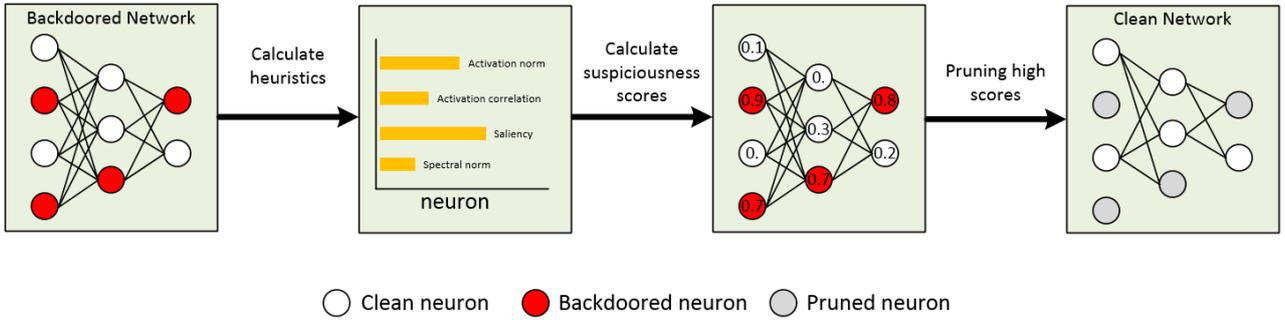

Figure 1. An overview of the proposed methodology for mitigating backdoors in a typical neural network.

Based on previous works' disadvantages, our proposed method concentrates on scenarios with limited data availability. Our approach is particularly applicable to cases where the defender lacks access to a vast set of clean samples, and traditional fine-tuning is unfeasible for the recovered model.

### III. PROPOSED METHOD

This section introduces the details of our proposed backdoor removal algorithm. An overview of the proposed method is shown in Figure 1. According to the figure, the proposed method calculates the suspiciousness score for each neuron or filter in the network based on four different heuristics. Thereafter, neurons with the highest suspiciousness score are pruned to access a clean network. Sections 3.1 and 3.2 describe how we consider each filter's sensitivity for the network's input and output. Afterward, in sections 3.3 and 3.4, we explain the intuition behind the activation norms and the correlation between poisoned and clean filters. Finally, in Section 3.5, we put everything together and propose a mathematical formula to calculate the backdoor suspiciousness score for each filter, which is used to find the poisoned filters in a CNN.

#### A. Spectral norm

This section explains spectral norm and how it helps locate poisoned neurons or filters in a deep network. Consider a matrix of transformation denoted as $T$, with dimensions $m \times n$. The operation of transformation, $T.v$, takes a vector $v$ in a space $\mathbb{R}^n$ and maps it onto a distinct vector within another space, $\mathbb{R}^m$. The decomposition of this transformation matrix using *singular value decomposition (SVD)* dissects $T$ into a sequence of three distinct geometric transformations: an orthogonal marix $V$, succeeded by a diagonal scaling matrix $\Sigma$, and further succeeded by another orthogonal matrix $U$. The matrix $\Sigma$ is diagonal in nature, and its diagonal elements correspond to the singular values of matrix $T$. In essence, when the singular values possess magnitudes greater than one, the transformation $T$ leads to the expansion of certain components of vector $v$, while concurrently contracting others when the singular values have magnitudes less than one [23].

In CNN, since each filter is associated with a weight matrix, the mentioned analysis on this matrix provides insights into how the filter intensifies or dampens fluctuations in its input space, as discussed in [21]. Regarding the weight matrix, the maximum singular value corresponds to the largest stretching factor along a specific direction in the input space of the associated filter, named the *spectral norm* [24]. When the maximum singular value is large, the filter is highly sensitive to input changes along this dominant direction. Consequently, slight variations in the input space caused by specific features or patterns can lead to significant changes in the response of the filter. In practice, filters with a high maximum singular value respond intensely to trigger patterns compared to neurons with a low maximum singular value [21]. Therefore, this characteristic can play a crucial role in locating poisoned neurons.

Moreover, the batch normalization (BN) [25] layer has been adopted in recent neural networks to stabilize the training process and make the optimization landscape much smoother. It normalizes the batch inputs of each channel and adjusts the mean and variance through trainable parameters. BN is usually placed after the convolution and before the activation function. Since BN records backdoor features when trained on a poisoned dataset [26], we merge convolutional layers with their corresponding BN to take the effect of BN statistics (with mean $\gamma$ and standard deviation $\sigma$) into consideration [21]. In conclusion, the spectral norm of a filter is calculated according to Equation 1.

$$spectral\ norm = max_{\Sigma_i} SVD(w \times \frac{\gamma}{\sigma}) \qquad (1)$$

In Equation 1, $w$ denotes the weights associated with a convolutional filter.

#### B. Saliency of neurons

As the second heuristic to detect poisoned neurons, we define our measurement of the contribution of each filter to the output of the network and explain how this contribution helps us in locating backdoored filters. *Gradient descent* is an iterative first-order optimization algorithm to find a given function's local minimum. This method is commonly used in machine learning and deep learning methods to minimize a loss function, given a dataset $D$, shown in Equation 2 [27].

$$min_w \frac{1}{|D|} \sum_{(x,y) \in D} l(w; x, y) \quad (2)$$

The derivative $\frac{\partial}{\partial x} f(t)$ measures how function $f$ changes as only the variable $x$ increases at point $t$. The "gradient" generalizes the notion of derivative to the case where the derivative is with respect to a vector, *i.e.*, multiple variables. The gradient of the loss function concerning the weights of a filter provides valuable information about how changes in the filter's output influence the overall loss function [28]. In other words, the gradient quantifies how much the loss function would change with a small perturbation in the filter's activation. Hence, it reasonably measures how sensitive the network's overall performance is to changes conducted by a filter. We take the average magnitude of the gradients of loss function w.r.t a filter, given by Equation 3.

$$saliency(n) = \sum_{(x,y) \in D} \frac{1}{|n|} \sum_{w \in n} |\frac{\partial}{\partial w} l(w; x, y)| \quad (3)$$

where, $D$ is the available defense set and $N$ denotes a set of the filters of the network.

### C. Activation norm

As the third heuristic, we explain how activations of filters guide us towards refining our suspiciousness score to effectively discriminate poisoned filters from clean ones. Convolutional networks require much fewer parameters to train than fully connected networks due to the assumption of locality, *i.e.*, only neighboring features should be processed together. Therefore, filters with rich parameters are highly correlated and express similar information to others in each layer [29]. This fact induced researchers to compress CNNs through *filter pruning* approaches, aiming to merge or remove filters representing similar features [30], [31].

Regarding activation of filters, the average norm of activation values of each filter is different for poisoned and clean samples, as shown in [32]. Hence, inspired by filter pruning approaches, we aim to prune those filters that are less activated by a set of clean samples. Our experimental results in this paper show filters with less activation are more prone to backdoor attacks since they may be taught to get activated by special features inserted by trigger or pattern on certain occasions. To calculate the activation of each filter, we take the average *l2-norm* of its output feature map, defined by Equation 4.

$$activation\ norm(n) = \frac{1}{|D|} \sum_{x \in D} \|activation_n(x)\|_2, \quad (4)$$
$$for\ all\ n \in N$$

where, $\|.\|_2$ denotes the *l2-norm* function.

### D. Correlation with other neurons

For the last heuristic, we define a measurement of how each filter is correlated with other filters inside each layer because these correlations give us a better insight into locating useless neurons.

Following the intuition of filter pruning approaches, we believe that filters with less activation correlation with other filters inside the same layer tend to have more potential to carry trojan features. Since we calculate the correlation according to the activation of filters on clean samples, it denotes a filter's degree of independence, and the more independent a filter is, the more potential it is to be poisoned. However, previous defense methods neglect the correlations between filters when pruning poisoned filters.

Given a set $N$ of all network filters, we calculate correlation for each filter according to equation 5.

$$correlation(i,j) = corrcoef\left(\|activation_i\|_2, \|activation_j\|_2\right) \quad (5)$$
$$for\ all\ i\ and\ j \in N$$

where, $\|.\|_2$ denotes the *l2-norm* function and $corrcoef(.)$ shows the correlation-coefficient function.

After calculating $correlation(.)$ for all filters of a layer, we normalize all correlation values within the range of *[0, 1]*.

### E. Putting everything together

Ultimately, through the synthesis of concepts clarified in preceding sections, we arrive at a measure of backdoor suspicion for each filter within a CNN. It is important to emphasize that our objective revolves around identifying filters characterized by minimal activation correlation and activation norm values while concurrently maximizing their saliency and spectral norm. To this aim, our focus is on discerning filters exhibiting the highest degree of backdoor suspicion, following the formulation presented in Equation 6.

$$suspiciousness\ score = \sqrt{\frac{saliency}{correlation \times activation\ norm}} \quad (6)$$
$$\times spectral\ norm$$

The inclusion of the square root function within an equation can be substantiated by its proficiency in skillfully adjusting the scale of numerical values associated with the coefficient of spectral norm in equation 6. The square root's intrinsic capacity to compress larger values while preserving relative differences offers potential enhancements in terms of numerical stability and the interpretability of outcomes.

The step-by-step process of the proposed method is detailed in Algorithm 1. The effectiveness of this method hinges on the careful selection of a threshold, *i.e.*, $\mu$. In this study, the threshold for the $l$-th layer is straightforwardly established as $\gamma_l + \mu \times \sigma_l$, where $\gamma_l$ and $\sigma_l$ represent the mean and standard deviation of suspiciousness scores in the $l$-th layer, respectively. The aforementioned threshold aligns with the conventional demarcation line used to identify outliers in a standard Gaussian distribution. Those filters with higher suspiciousness scores above the threshold are considered backdoored and further pruned.

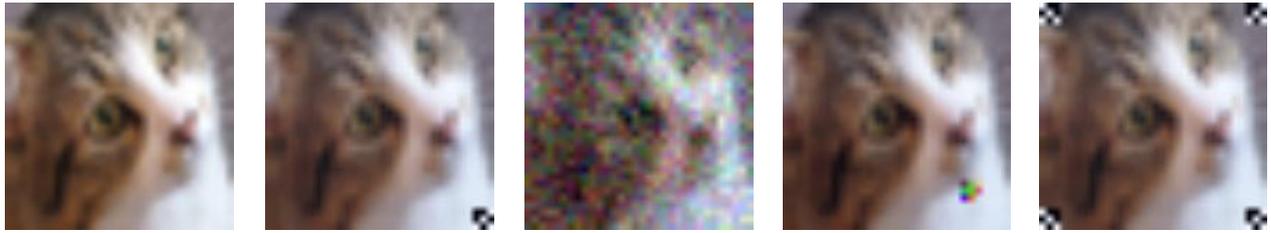

(a) Original    (b) BadNets    (c) Blend    (d) Trojan    (e) CL

Figure 2. Illustrations of poisoned samples with different backdoor attacks.

As discussed in Section 3.1, the batch normalization layer (BN) records the statistics of backdoors when the network is trained on poisoned samples [26]. The rationale for prioritizing the pruning of BN neurons is inspired by ANP [19], which acknowledges that BN parameters may recover trojan features. Hence, pruning BN neurons in line 6 of Algorithm 1 disables both corresponding convolutional filters and BN statistics simultaneously, resulting in more robust backdoor mitigation.

---
**Algorithm 1. Backdoored Filter Pruning**

**Input:** A set of layers including all filters; threshold hyper-parameter $\mu$
**Output:** describe the output of the algorithm
1. **for** layer $l \in$ layers **do**
2.     **for** filter $f \in$ filters of $l$ **do**
3.         Calculate spectral norm, saliency, activation norm, and activation correlation of $f$
4.         Calculate the suspiciousness score of $f$ using Equation 6
5.     Locate filters with scores higher than $\gamma_l + \mu \times \sigma_l$
6.     Prune BN neurons associated with determined filters

---

## IV. EXPERIMENTS

In this section, we conduct comprehensive experiments to evaluate the effectiveness of the proposed algorithm.

### A. Experimental settings

This section provides the details of attacks and baseline methods' configurations. Moreover, we define the metrics used to compare different defense methods in terms of the performance of the recovered model on clean and poisoned data.

#### A.1 Attack settings

We apply our proposed algorithm to a range of typical attack methods, namely BadNets [33], Blended Backdoor Attack [7], Trojan-square Attack [9], and Clean Label (CL) Attack [34]. We depict the poisoned samples and the adversarial instances generated by different adversarial techniques in Figure 2. Similar to prior research, we assess the effectiveness of our defense against these four attacks on the CIFAR-10 dataset using the ResNet-18 architecture. We follow the recommended default configurations outlined in their original publications and employ the provided open-source codes for all attacks, including trigger patterns and hyper-parameters. The dimensions of the trigger used in our experiments are set at $3 \times 3$. In addition, we establish the target label for all attacks as the 0-th class, and the contamination rate is fixed at 1%. Moreover, all experiments are conducted within the PyTorch framework.

#### A.2 Baseline methods

We compare our algorithm with the most advanced backdoor defense methods available, which include Neural Attention Distillation (NAD) [18], Adversarial Neuron Pruning (ANP) [19], Channel Lipschitzness Pruning (CLP) [21], and Adversarial Weight Masking (AWM) [20], in addition to our proposed algorithm. All these defenses rely on a limited set of clean samples, *i.e.*, just one sample per label, for their defense mechanism. More specifically, the defense set is assumed to contain only ten image samples for the CIFAR-10 dataset. We refer to the open-source implementations of NAD, ANP, CLP, and AWM, making necessary adjustments to their hyper-parameters to achieve optimal performance against various attacks. As for the proposed algorithm, we determine the value of the hyper-parameter $\mu$ through experiments conducted in Section 4.2.2.

### B. Evaluation metrics

We employ two metrics to assess the effectiveness of our defense strategy: (1) Clean Accuracy (ACC), which gauges how accurately the model performs on clean test data, and (2) Attack Success Rate (ASR), which indicates the model's accuracy when dealing with poisoned test data. It is important to note that ASR signifies the proportion of poisoned samples mistakenly classified as the intended target label. This rate is calculated using the backdoored samples whose actual labels are not associated with the target attack category. An effective defense mechanism should achieve a low ASR while preserving the ACC at a high level.

Table I Performance assessment of the proposed algorithm and four SOTA defense methods using a set of 10 clean samples for the CIFAR-10 dataset.

| Attack Type | | Backdoored | NAD | ANP | AWM | CLP | Ours |
|---|---|---|---|---|---|---|---|
| BadNets | ASR | 100 | 2.72 | <u>2.28</u> | 64.98 | 7.15 | **1.27** |
| | ACC | 93.39 | 82.87 | 80.11 | 74.61 | 86.60 | **89.23** |
| Blend | ASR | 99.92 | 91.39 | 100 | 6.92 | <u>3.0</u> | **1.38** |
| | ACC | 93.22 | 80.81 | 77.32 | 83.67 | 90.25 | **90.59** |
| Trojan | ASR | 99.90 | 84.47 | <u>8.92</u> | 9.34 | 11.37 | **4.67** |
| | ACC | 93.22 | 81.81 | 68.96 | 75.54 | 82.14 | **86.19** |
| CL | ASR | 99.94 | 32.58 | 89.76 | 75.26 | <u>4.54</u> | **2.28** |
| | ACC | 93.78 | 49.69 | 88.30 | 72.65 | **90.98** | 90.31 |
| **Average** | ASR | 99.94 | 52.79 | 50.24 | 39.12 | <u>6.51</u> | **2.4** |
| | ACC | 93.40 | 73.79 | 78.67 | 76.61 | <u>87.49</u> | **89.08** |

## C. Experimental results

In this section, we analyze the performance of the proposed defense algorithm in comparison to baselines. Then, we provide more experiments on the effect of hyper-parameter μ, an ablation study of various parts of the formulation provided in previous sections, and the running time of the proposed algorithm against baselines. In order to structure the results of this section, we pose five research questions as following,

- RQ1 (Effectiveness): How effective is the proposed method against common backdoor attacks?
- RQ2 (Comparison): How does the proposed method compare to other defense methods?
- RQ3 (Hyper-parameter tuning): What impact does the hyper-parameter $\mu$ have on the proposed algorithm?
- RQ4 (Heuristics effectiveness): How effective are the heuristics separately in comparison to Equation 6?
- RQ5 (Efficiency): How efficient is the proposed method in terms of running time?

These research questions will be answered through this section and various experiments.

*RQ1 (Effectiveness).* How effective is the proposed method against common backdoor attacks?

To evaluate the effectiveness of the proposed method, we designed experiments against four common backdoor attacks. The results shown in the last column of Table I show that the proposed method achieves an ASR rate of 2.4% and ACC rate of 89.08% on average. These evaluations encompass four prevalent attacks on the CIFAR-10 dataset utilizing the ResNet-18 architecture. Notably, superior ACC and reduced ASR are deemed favorable outcomes. The most optimal results are highlighted in bold, and the second-lowest ASR is underscored.

It is worth mentioning that each defense method has limitations against specific backdoor attacks. For example, although our proposed method has the best overall performance, there is still room for improving its performance against Trojan attack in terms of both ASR and ACC. Exploring ways to improve the performance of the proposed method against Trojan attacks is an area of future work for our research.

*Finding for RQ1.* Our proposed method is effective in mitigating the backdoored network. Our evaluations prove that the proposed method repairs the neural backdoor by pruning neurons which reduces successful attack likelihood by 97.5% on average.

*RQ2 (Comparison).* How does the proposed method compare to other defense methods?

The results displayed in Table I reveal that the proposed algorithm consistently surpasses the fundamental methods in numerous scenarios, thereby showcasing its effectiveness. Compared to NAD, ANP, and AWP, a notable advantage of our algorithm is its reduced reliance on extensive clean data, making it more versatile for application in domains where obtaining a substantial set of clean data is challenging. Our approach also demonstrates superior performance in terms of backdoor resilience due to its utilization of data-specific characteristics compared to CLP. On the other hand, while CLP achieves considerable results against most attacks, it has weaker performance against BadNets attacks compared to ANP, which reduces the ASR by 97.72%.

Table II The statistical test compares the performance of the proposed method with other baselines.

| Attack Type | | Wilcoxon p-values | | | |
|---|---|---|---|---|---|
| | | NAD | ANP | AWM | CLP |
| BadNets | ASR (-+++) | 0.9677 | <0.001 | 0.0029 | 0.0419 |
| | ACC (++++) | <0.001 | <0.001 | 0.0019 | <0.001 |
| Blend | ASR (-+++) | 0.9677 | 0.0048 | 0.0185 | 0.0185 |
| | ACC (++++) | <0.001 | <0.001 | <0.001 | 0.0019 |
| Trojan | ASR (++++) | <0.001 | 0.0185 | <0.001 | 0.0029 |
| | ACC (-+++) | 1.0 | <0.001 | <0.001 | <0.001 |
| CL | ASR (-+++) | 0.9677 | <0.001 | <0.001 | 0.0419 |
| | ACC (+++-) | <0.001 | <0.001 | <0.001 | 0.9345 |

Evidently, in the case of CLP defending against CL attacks, there is a noticeable increase in ACC. However, this method comes at the expense of a higher ratio of ASR, indicating a tradeoff between accuracy and vulnerability. In contrast, our proposed approach strikes a harmonious equilibrium between ASR and ACC, effectively avoiding such tradeoffs. Although CLP demonstrates commendable outcomes against the majority of attacks without necessitating data, it falls short in accurately identifying the backdoor neurons due to its exclusive focus on input sensitivity. Regarding ASR, our approach exceeds the performance of all baseline methods by at least about 50% when countering BadNets, Blend, Trojan, and CL attacks. Regarding ACC, our method outperforms baseline methods by 3-7% degradation. In the context of other attack scenarios, our method achieves nearly equivalent ACC compared to the second-highest-performing competitors.

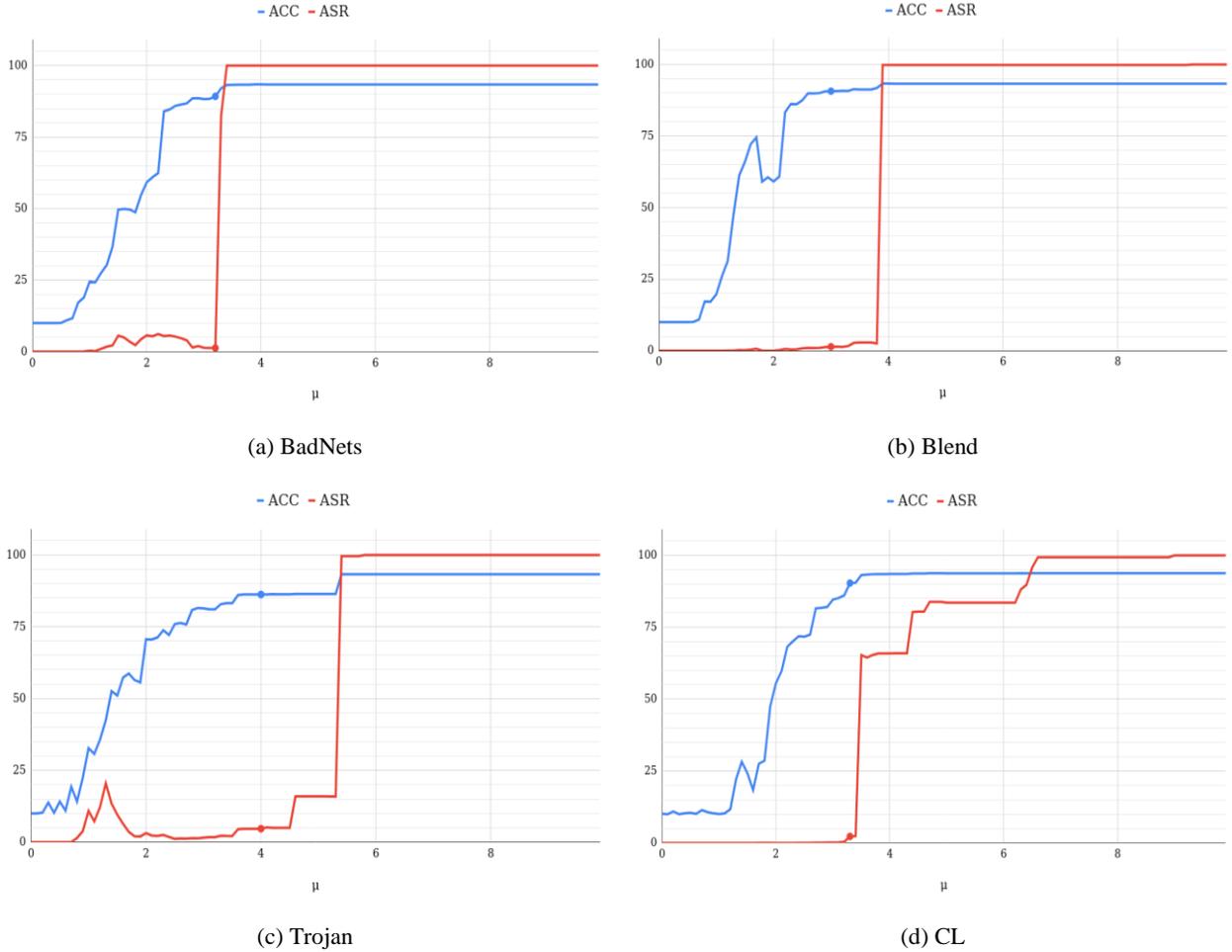

Figure 3. Performance of the proposed algorithm with different values for hyper-parameter $\mu$ against different attacks on CIFAR-10 with ResNet-18.

Regarding statistical evaluation, we validate the superior performance of the proposed method using the Wilcoxon signed-rank test [35] for statistical significance at a 95% confidence level for ten independent runs. Wilcoxon singed-rank test is a non-parametric statistical hypothesis test which compares samples to measure whether the mean of their underlying distribution differ. Moreover, each run of the algorithm chooses a different set of images to mitigate the backdoor. Our null hypothesis of the Wilcoxon test is that the mean of the underlying distribution of metrics ASR and ACC, is the same as the baselines. As the alternative hypothesis, we assume that the mean of the metrics of the proposed method is higher than the baselines. The results shown in Table II reveal that the proposed method significantly ($p$-value $< 0.05$) outperform the baselines. Notably, superior p-values are marked by "+" sign. According to the results, the proposed method demonstrate better ASR rate in 81.25% of experiments, while it reaches 87.5% in terms of ACC metric.

*Finding for RQ2.* The proposed method statistically outperforms the baselines, with 81.25% and 87.5% likelihood according to ASR and ACC metrics, respectively.

*RQ3 (Hyper-parameter tuning).* What impact does the hyper-parameter $\mu$ have on the proposed algorithm?

As detailed in Section 3.5, the hyper-parameter $\mu$ governs the balance between clean data test accuracy and resilience against backdoor attacks. Figure 3 shows the ACC and ASR employing varying hyperparameter values for $\mu$. Note that the best values of the threshold are marked using a circle in both curves.

Notably, we observe a swift decline in ASR to nearly 0% as $\mu$ decreases, while ACC experiences a slower decline. This behavior indicates that channels associated with backdoor activity tend to possess higher suspiciousness scores than regular channels, a characteristic that our method accurately prunes. It is important to emphasize that as $\mu$ approaches 0, a substantial proportion of channels in the model are pruned, potentially leading to irrational model predictions. This accounts for the rapid decrease in ASR and ACC curves for some attacks when $\mu$ reaches 0.

Table III Comparison results of each part of the proposed formulation for backdoor suspiciousness.

| Attack Type | | Backdoored | Spectral norm | Saliency | Activation | Correlation | Equation 6 (without Sqrt) | Equation 6 |
|---|---|---|---|---|---|---|---|---|
| BadNets | ASR | 100 | 7.15 | 89 | 100 | 1.18 | 7.82 | 1.27 |
|  | ACC | 93.39 | 86.60 | 75.26 | 90.93 | 69.97 | 91.39 | 89.23 |
| Blend | ASR | 99.92 | 3.0 | 16.61 | 99.90 | 0.10 | 1.06 | 1.38 |
|  | ACC | 93.22 | 90.25 | 82.57 | 92.01 | 83.14 | 88.93 | 90.59 |
| Trojan | ASR | 99.90 | 11.37 | 56.54 | 75.02 | 85.54 | 1.35 | 4.67 |
|  | ACC | 93.22 | 82.14 | 61.75 | 92.50 | 35.50 | 83.19 | 86.19 |
| CL | ASR | 99.94 | 4.54 | 35.28 | 55.95 | 7.38 | 0.73 | 2.28 |
|  | ACC | 93.78 | 90.98 | 75.20 | 93.51 | 84.85 | 89.77 | 90.31 |
| **Average** | ASR | 99.94 | 6.51 | 48.60 | 48.25 | 23.75 | 2.74 | 2.4 |
|  | ACC | 93.40 | 87.49 | 73.69 | 92.23 | 68.36 | 88.32 | 89.08 |

Conversely, $\mu$ values of greater or equal to 4 in cases of BadNets and Blend attacks and greater or equal to 8 in defense against Trojan and Clean-Label attacks, both the ASR and ACC metrics are equal to the ASR and ACC metrics of the original poisoned model. This behavior comes from the fact that higher threshold values prune none of the poisoned filters, and hence, no change is imposed to the compromised network.

Nevertheless, the optimal values for hyper-parameter $\mu$ for various attacks are different. For instance, a value of $\mu = 3$ turns out to be the optimal choice in the experiment against the Blend attack. However, this choice will not work properly for other attacks. In general, a value between 3 and 4 generalizes well on different attacks, according to Figure 3.

> *Finding for RQ3.* A value between 3 and 4 turns out to be optimal for the threshold $\mu$ and makes the porposed algorithm generalize better to various attacks. Add to this, Increasing the threshold $\mu$ to larger values may end up the original poisoned model, due to ignoring poisoned filters.

> *RQ4 (Heuristics effectiveness).* How effective are the heuristics separately in comparison to Equation 6?

In this RQ, we perform an ablation study to verify the effectiveness of our proposed backdoor suspiciousness score formulation. The evaluation metrics used in this section are the same as in Table I, *i.e.*, ACC and ASR. All formulation variants are evaluated on the same backdoored model and clean dataset. The baseline methods consider each element individually to evaluate its effectiveness in pruning the most responsible filters for trojan behavior. Specifically, we separately prune filters based on spectral norm, saliency value, activation norm, and activation correlation.

An experiment is also conducted to investigate the impact of the square root operator on our final suspiciousness scoring equation. The results are shown in the last two columns of Table III. It is observed that the square root is crucial for the original formulation and leads to better results, *i.e.*, reducing the ASR by 0.3% and increasing ACC by almost 1% on average.

The performance results reported in Table III demonstrate that, although each element is effective on its own, combining them along with the scaling square root operator locates the most responsible filters to backdoor behavior and achieves better results regarding ACC and ASR. On average, Equation 6 achieves an ASR rate of 2.4% and an ACC rate of 89.08, which is superb compared to single heuristics.

> *Finding for RQ4.* Equation 6 outperforms all separate heuristics in terms of both ASR and ACC metrics. The scaling square root operator is helpful in imrpoving the performance of Equation 6 to score sispicious filters and neurons.

> *RQ5 (Efficiency).* How efficient is the proposed method in terms of running time?

We have recorded the execution times of the aforementioned defense strategies using ResNet-18 on CIFAR-10 images, and the outcomes are documented in Table IV. All methods except CLP were assessed on the Google CoLab platform with a Tesla K80 GPU. On the other hand, CLP was executed on a CPU, specifically the 11th Gen Intel Core i7-1165G7 @ 2.80GHz. Notably, our proposed algorithm outpaces NAD, ANP, and AWM in terms of processing speed. This is attributable to our algorithm's requirement of only one forward passing the small defense set through the neural network. Remarkably, CLP emerges as the swiftest approach due to its data-independent nature. The reported running times encompass the time taken for the fine-tuning procedure in NAD, the process of adversarial perturbation and mask learning in ANP, and the mask learning process in AWM.

Table IV The overall running time of different defense methods on CIFAR-10 images with ResNet-18. All the methods except for CLP are trained on GPU.

| Defense method | NAD | ANP | AWM | CLP | Ours |
|---|---|---|---|---|---|
| Running time (sec.) | 170.6 | 666.83 | 432.36 | 35.06 | 48.92 |

> ***Finding for RQ5.*** The time overhead of the proposed algorithm is reduced by more than 80% compared to the fastest baseline consuming data, *i.e.*, NAD.

## V. CONCLUSION

In this paper, we formulate backdoor suspiciousness of poisoned neurons in a compromised DNN, which characterizes their activation and correlation with other neurons, sensitivity to input, and contribution to the network output. Based on these attributes, we propose an algorithm to prune neurons with suspiciousness scores higher than a threshold. Experiments show that the proposed method achieves more than 50% accuracy in detecting and pruning backdoored poisoned neurons, even with a tiny dataset of clean samples. Moreover, our proposed backdoor attack repair method is three times faster than the state-of-the-art neural trojan repair methods.

For future work, we intend to investigate an ensemble of pruned models. Each pruned model is a cleaned network with respect to each heuristic used to score neurons' suspiciousness. Eventually, the predictions of all experts are aggregated to reach a final decision for the corresponding class of the input data.

## ACKNOWLEDGEMENTS

During the preparation of this work the authors used OpenAI ChatGPT-3.5 in order to improve the readability of the manuscript. After using this tool/service, the authors reviewed and edited the content as needed and take full responsibility for the content of the publication.

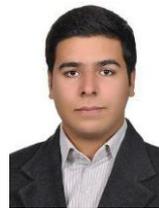

**Soroush Hashemifar** received his B.Sc. degree in computer engineering from Razi University, and his M.Sc. degree in software engineering from Iran University of Science and Technology in 2020 and 2023, repsectively. His research interests include deep neural networks, explainble AI, AI safety, and software engineering.

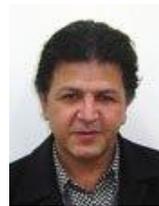

**Saeed Parsa** is an Associate Professor in the Software Department of the School of Computer Engineering and the Director of the Reverse Engineering Research Laboratory (formerly parallel processing laboratory) at Iran University of Science and Technology. His research interests are in software testing, automated software engineering, compilers, and reverse software engineering. At present, the projects of their master and doctoral students are on the topics of automatic production of test data, statistical location of hidden errors, fuzzy test and automatic repair of programs.

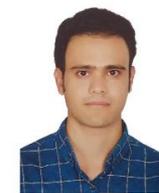

**Morteza Zakeri-Nasrabadi** received his M.Sc. degree in Software Engineering as the first rank student from the Iran University of Science and Technology (IUST) in 2018. He is a Ph.D. student in the School of Computer Engineering at Iran University of Science and Technology, Iran, Tehran. Morteza does research in Software Engineering and Artificial Intelligence. His research interests are automated and intelligent software engineering (AISE), especially automated software refactoring, testing, program analysis, and applying machine learning in software engineering. Currently, he works on applying machine learning techniques to software engineering, mainly code-related tasks, such as software refactoring, testing, and quality measurement.